\documentclass[11pt]{article}
\usepackage{acl}

\usepackage{times}
\usepackage{latexsym}

\usepackage[T1]{fontenc}
\usepackage[utf8]{inputenc}
\usepackage{microtype}
\usepackage{inconsolata}

\usepackage{amsmath}
\usepackage{amssymb}
\usepackage{graphicx}
\graphicspath{{./}{../}}
\usepackage{booktabs}
\usepackage{multirow}
\usepackage{makecell}
\usepackage{float}
\usepackage{subcaption}
\usepackage{enumitem}
\usepackage[table]{xcolor}
\usepackage{colortbl}
\usepackage{tcolorbox}
\usepackage{listings}
\usepackage{dblfloatfix}
\tcbuselibrary{skins, breakable}

\definecolor{MethodOrange}{RGB}{255, 225, 190}
\definecolor{DataYellow}{RGB}{255, 252, 225}
\definecolor{MethodPink}{RGB}{255, 215, 225}
\definecolor{DataPink}{RGB}{255, 240, 245}
\definecolor{TextOrange}{RGB}{210, 105, 30}
\definecolor{TextPink}{RGB}{200, 20, 100}
\definecolor{papergreen}{HTML}{008000}
\definecolor{paperred}{HTML}{D00000}
\definecolor{seedblue}{HTML}{2E5AA8}
\definecolor{headerBlue}{RGB}{68, 114, 196}
\definecolor{bgBlue}{RGB}{217, 226, 243}
\definecolor{headerGreen}{RGB}{84, 130, 53}
\definecolor{bgGreen}{RGB}{226, 239, 218}
\definecolor{decisionRed}{RGB}{192, 0, 0}
\definecolor{taskBlue}{HTML}{2F6FF2}
\definecolor{taskPink}{HTML}{F04B78}
\definecolor{taskYellow}{HTML}{D98A00}

\definecolor{baseRow}{gray}{0.96}
\definecolor{dpoRow}{HTML}{F8DDE8}
\definecolor{gain}{RGB}{0,122,75}
\definecolor{loss}{RGB}{200,45,45}
\definecolor{deltaz}{gray}{0.55}
\newcommand{\baserow}{\rowcolor{baseRow}}
\newcommand{\dporow}{\rowcolor{dpoRow}}
\newcommand{\dplus}[1]{\textcolor{gain}{\scriptsize\,#1}}
\newcommand{\dminus}[1]{\textcolor{loss}{\scriptsize\,#1}}

\newlist{rQList}{enumerate}{1}
\setlist[rQList]{
    label=\textbf{\textbullet~RQ\arabic*:},
    ref=RQ\arabic*,
    leftmargin=1.6em,
    labelsep=0.5em,
    align=left,
    nosep
}

\title{NE-R1: Enhancing Named Entity Recognition Model via Reinforcement Learning}

\author{
\textbf{Meixuan Chen\textsuperscript{1}}
\thanks{Equal contribution.\quad
$^{\dagger}$~Corresponding author.\protect\newline
$^{\ddagger}$~Work done during an internship at Baidu, Inc.}
\footnotemark[3],
\textbf{Hehan Li\textsuperscript{2}}\footnotemark[1],
\textbf{Ruizhi Zhao\textsuperscript{3}},
\textbf{Xin Lu\textsuperscript{2}},
\textbf{Peizhi Xu\textsuperscript{2}},
\\
\textbf{Liwei Qian\textsuperscript{2}},
\textbf{Meifang Li\textsuperscript{2}},
\textbf{Shuanglong Li\textsuperscript{2}},
\textbf{Hanmeng Liu\textsuperscript{2}},
\\
\textbf{Xin Pei\textsuperscript{2}},
\textbf{Yanbiao Ma\textsuperscript{4,5,6}}\footnotemark[2]
\\[2mm]
\textsuperscript{1}Shenzhen Graduate School, Peking University
\\
\textsuperscript{2}Baidu, Inc., Beijing, China
\\
\textsuperscript{3}Shenzhen University, Shenzhen, China
\\
\textsuperscript{4}Gaoling School of Artificial Intelligence,
Renmin University of China, Beijing, China
\\
\textsuperscript{5}Beijing Key Laboratory of Research on
Large Models and Intelligent Governance
\\
\textsuperscript{6}Engineering Research Center of
Next-Generation Intelligent Search and Recommendation, MOE
\\[1mm]
\small{
\texttt{chenmeixuan26@stu.pku.edu.cn}
\quad
\texttt{2410145008@mails.szu.edu.cn}
\quad
\texttt{ybma1998@ruc.edu.cn}
}
\\
\small{
\texttt{\{lihehan,luxin06,xupeizhi,qianliwei,limeifang,lishuanglong,liuhanmeng\}@baidu.com}
}
}

\begin{document}

\maketitle

\begin{abstract}
Named Entity Recognition (NER) has achieved substantial progress since the advent of large language models (LLMs). Nevertheless, the recognition of long-tail and domain-specific entities remains challenging due to the deficiency in parametric knowledge. Retrieval-augmented generation (RAG) offers a promising remedy by injecting external knowledge, but it also introduces noise and unnecessary cost when dealing with familiar cases. In this paper, we propose \textbf{NE-R1}, a novel framework for adaptive retrieval-augmented NER. We design a  ``retrieval-on-demand'' mechanism for NER. Then we integrate it into models by a two-stage training method: (1) multi-task instruction tuning initialization; (2) end-to-end RL optimization with CoT. To achieve reasonable selection between parameterized and external knowledge, we design a multi-dimensional reward considering both accuracy and retrieval benefit. NE-R1 achieves state-of-the-art performance on various benchmarks,
with average improvements of  \textbf{2.52} F1 points in in-domain evaluation and \textbf{1.18} F1 points in zero-shot cross-domain evaluation.
\end{abstract}

\section{Introduction}
Named Entity Recognition (NER) is a fundamental task in Natural Language Processing (NLP), aiming to identify entity spans and their types \cite{lample2016neural, ma2016end}. In recent years, large language models (LLMs) have catalyzed the rapid advancement of Generative NER (GenNER) \cite{seow2025review, wang2023instructuie, zhou2024universalner}, which reformulates NER as a conditional generation problem, and achieved remarkable progress in this area.

\begin{figure}[!t]
    \centering
    \includegraphics[width=1.01\columnwidth]{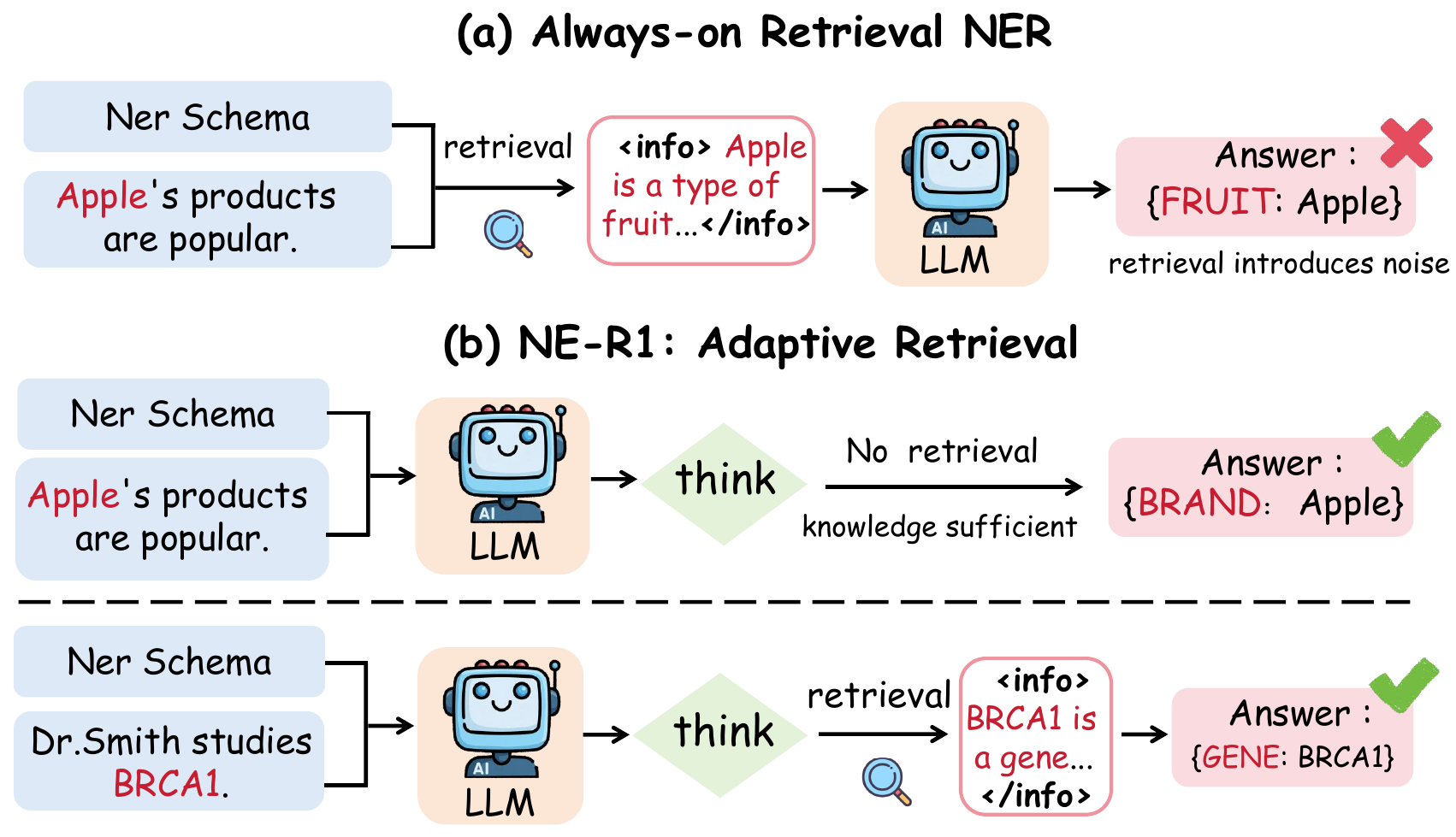}
    \vskip -0.1in
    \caption{Motivation and overview of NE-R1's adaptive retrieval mechanism. (a) Always-on retrieval indiscriminately introduces redundant noise and degrades performance for familiar entities. (b) NE-R1 adaptively triggers retrieval only when genuinely necessary.}
    \label{fig:trigger}
\end{figure}

However, GenNER still suffers from the parametric knowledge deficiency of LLMs \cite{lu2022unified, guo2025baner}. This leads to hallucinations or omissions when encountering long-tail, domain-specific unfamiliar entities or ambiguous situations \cite{kang2025unfamiliar}. To address this limitation, Retrieval-Augmented Generation (RAG) \cite{gao2023retrieval, fan2024survey} has been introduced to inject external knowledge into NER models \cite{tan2023damo, zhenwei2024ra}, and further augmentation has been explored through retrieving similar examples \cite{li2024wkner}.

Despite the benefits brought by retrieval augmentation, existing methods predominantly adopt a ``\textbf{full retrieval}'' strategy---triggering retrieval indiscriminately regardless of input complexity. This paradigm engenders two critical issues: \textbf{(1) Noise Introduction}, where mandatory retrieval for familiar cases may inject irrelevant information that interferes with model decisions, as illustrated in Figure~\ref{fig:trigger}(a). \textbf{(2) Unnecessary Latency Overhead}, where retrieval for high-frequency entities is often redundant. Our experiments on the MIT-Movie dataset reveal that full retrieval incurs a $4.85\times$ increase in inference latency compared to no retrieval, yet yields negligible performance gains. Consequently, it becomes paramount to realize an adaptive ``\textbf{retrieve-on-demand}'' mechanism and seamlessly integrate it into NER models.

In this paper, we propose \textbf{NE-R1}, a novel framework for adaptive retrieval-augmented NER. As shown in Figure~\ref{fig:trigger}(b), NE-R1 intelligently determines when to trigger retrieval based on input complexity. Inspired by the recent breakthroughs of DeepSeek-R1 \cite{guo2025deepseekr1} in reasoning tasks, we design a two-stage training procedure. The first stage is capability initialization through multi-task instruction tuning, which endows the model with three foundational capabilities: parametric inference, retrieval triggering, and evidence fusion; the second stage is end-to-end RL training, incorporating an explicit Chain-of-Thought mechanism and a multi-dimensional reward that comprehensively estimates prediction correctness and retrieval benefit to jointly enhance overall performance.

Extensive experiments show that NE-R1 outperforms the evaluated baselines, improving the average in-domain F1 score by \textbf{2.52} points over the strongest overall baseline, including a \textbf{3.30}-point gain on GENIA. In zero-shot cross-domain evaluation, NE-R1 improves performance in four of the five CrossNER domains and achieves the best five-domain average, exceeding the strongest baseline by \textbf{1.18} F1 points. On MIT-Movie, NE-R1 also reduces inference latency by
approximately \textbf{55\%} compared with the always-retrieve baseline while achieving a higher F1 score. Our contributions are summarized as follows:

\begin{itemize}[left=0pt, itemindent=1em]
    \item We propose NE-R1, a novel framework that brings a ``retrieve-on-demand'' mechanism to NER, allowing models to retrieve only when their own knowledge falls short.
    \item We design a two-stage training method that first injects foundational capabilities via multi-task instruction tuning, then performs end-to-end RL optimization with Chain-of-Thought reasoning and a multi-dimensional reward.
    \item NE-R1 surpasses existing methods on both in-domain and zero-shot cross-domain evaluations, demonstrating an  favorable balance of performance, efficiency, and generalization.
\end{itemize}

\begin{figure*}[t]
    \centering
\includegraphics[width=1\textwidth]{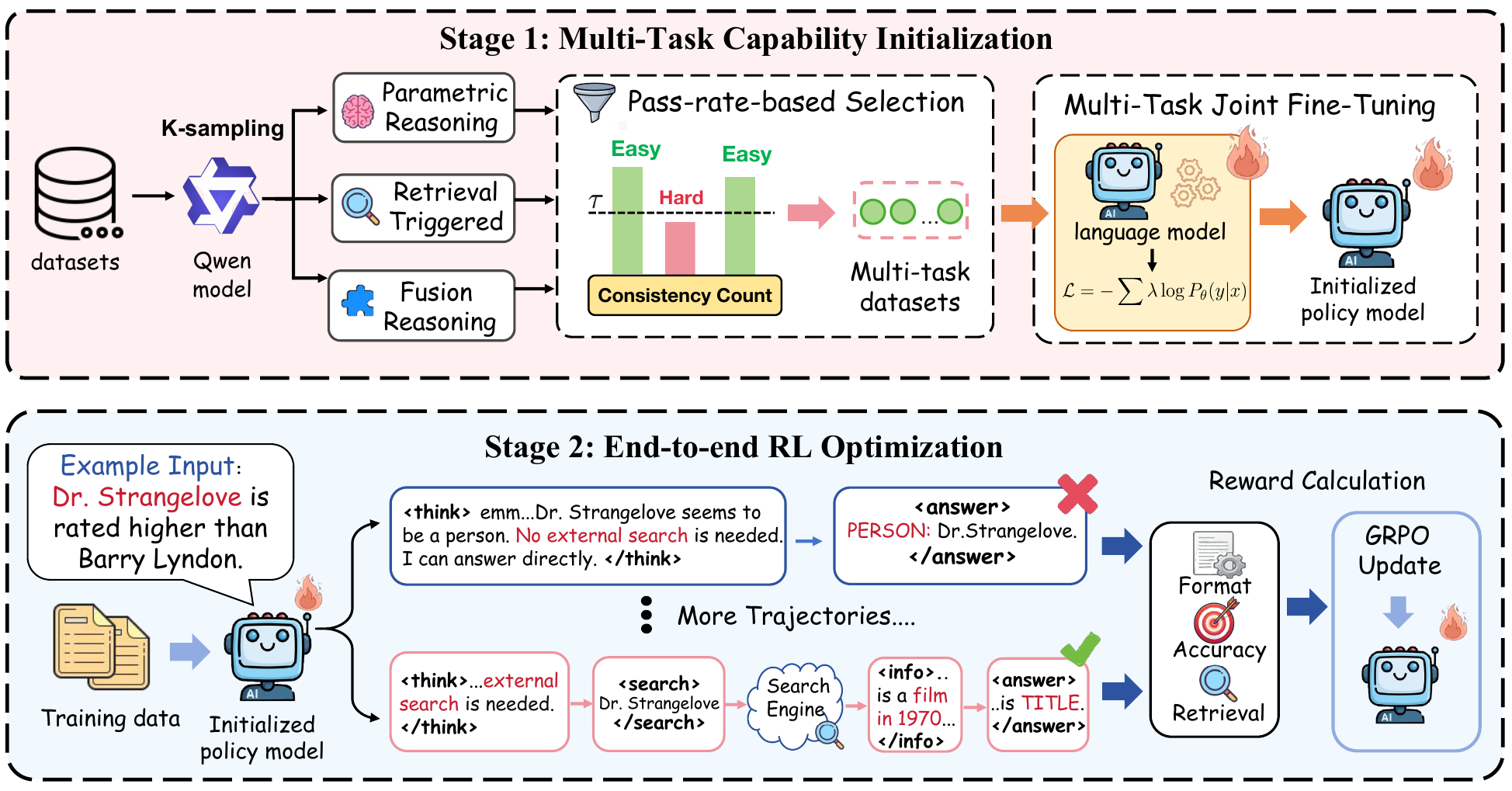}
    \caption{Overview of the \textsc{NE-R1} framework. The pipeline consists of two main stages: (1) \textbf{Multi-Task Capability Initialization}, where we construct three instruction tuning tasks via pass-rate-based data selection to equip the model with foundational capabilities for parametric inference, retrieval triggering, and fusion inference; and (2) \textbf{End-to-End RL Optimization}, which employs Group Relative Policy Optimization (GRPO) to fine-tune the model, leveraging Chain-of-Thought (CoT) guided reasoning and a multi-dimensional reward mechanism to achieve adaptive retrieval and precise entity extraction.}
    \label{fig:framework}
\end{figure*}

\section{Related Work}

\paragraph{LLM-based NER.} Recent LLM-based NER work primarily follows two main paradigms. The first is In-Context Learning \cite{dong2024survey}, which directly invokes LLMs through prompts. Works in this direction focus on instruction optimization \cite{liu2023pre, wang2025novel, gonen2023demystifying}, providing domain knowledge \cite{tong2025evoprompt, cocchieri2025zeroner, huang2025guidener}, or introducing reflection mechanisms \cite{ashok2023promptner, xie2023empirical}. ICL example selection \cite{xie2024self, chen2025allabel} is also explored.

The second paradigm fine-tunes pretrained LLMs with supervised training. Representative works employ unified instruction templates \cite{lu2022unified}, incorporate Chain-of-Thought \cite{huang2026reasoning}, or decompose NER into subtasks \cite{guo2025baner, lu2024padellm}. For domain adaptation, methods leverage data augmentation \cite{huang2025adversity, bogdanov2024nuner}, inject guiding knowledge \cite{yang2025beyond, sainz2024gollie}, or denoise pseudo-labels \cite{ding2024improving}.

\paragraph{Retrieval-Augmented Generation.} RAG integrates external information to supplement knowledge and reduce hallucination \cite{zhao2026retrieval, peng2025graph, ram2023context}. Recent advances optimize retrieval methods \cite{yan2024corrective, gao2023precise, jeong2024adaptive}, perform document refinement \cite{trivedi2023interleaving, liu2026truthfulrag}, or design multi-round mechanisms \cite{shao2023enhancing}. However, most RAG systems rely on static retrieval strategies that may not adapt to diverse query types.

\paragraph{Reinforcement Learning for LLMs.} Reinforcement learning has been applied to LLMs\cite{ouyang2022training}, though early methods face tremendous training challenges \cite{schulman2017proximal}. Alternative approaches maximize reward by solving a classification problem\cite{rafailov2023direct} or optimize through advantage normalization \cite{hu2025reinforce++}. RL has been extended to machine translation \cite{he2025r1, feng2025mt}, text-to-SQL \cite{pourreza2025reasoning}, and question-answering \cite{song2025r1searcherplusplus, zheng2025deepresearcher}.

\section{Methodology}
In this section, we present the NE-R1 framework. As shown in Figure~\ref{fig:framework}, it comprises two key components: (1) \textbf{Multi-Task Capability Initialization (MTCI)}, which equips the model with capabilities in parametric inference, retrieval triggering, and evidence-fusion inference via multi-task instruction tuning (Section~\ref{sec:mtai}); and (2) \textbf{End-to-End RL Optimization}, which designs a Chain-of-Thought (CoT)-guided reasoning paradigm and a multi-dimensional reward, and employs Group Relative Policy Optimization (GRPO) for end-to-end optimization (Section~\ref{sec:rlpo}).

\subsection{Multi-Task Capability Initialization}
\label{sec:mtai}
We aim to enable adaptive retrieval decisions for NER. However, uninitialized models cannot generate effective queries or integrate retrieved results, making direct reinforcement learning inefficient and unstable. We therefore apply multi-task instruction tuning to inject three foundational capabilities into Qwen2.5-7B, using supervised data distilled from a stronger teacher model, Qwen2.5-32B.

\begin{figure*}[t]
    \centering
    \includegraphics[width=1.0\textwidth]{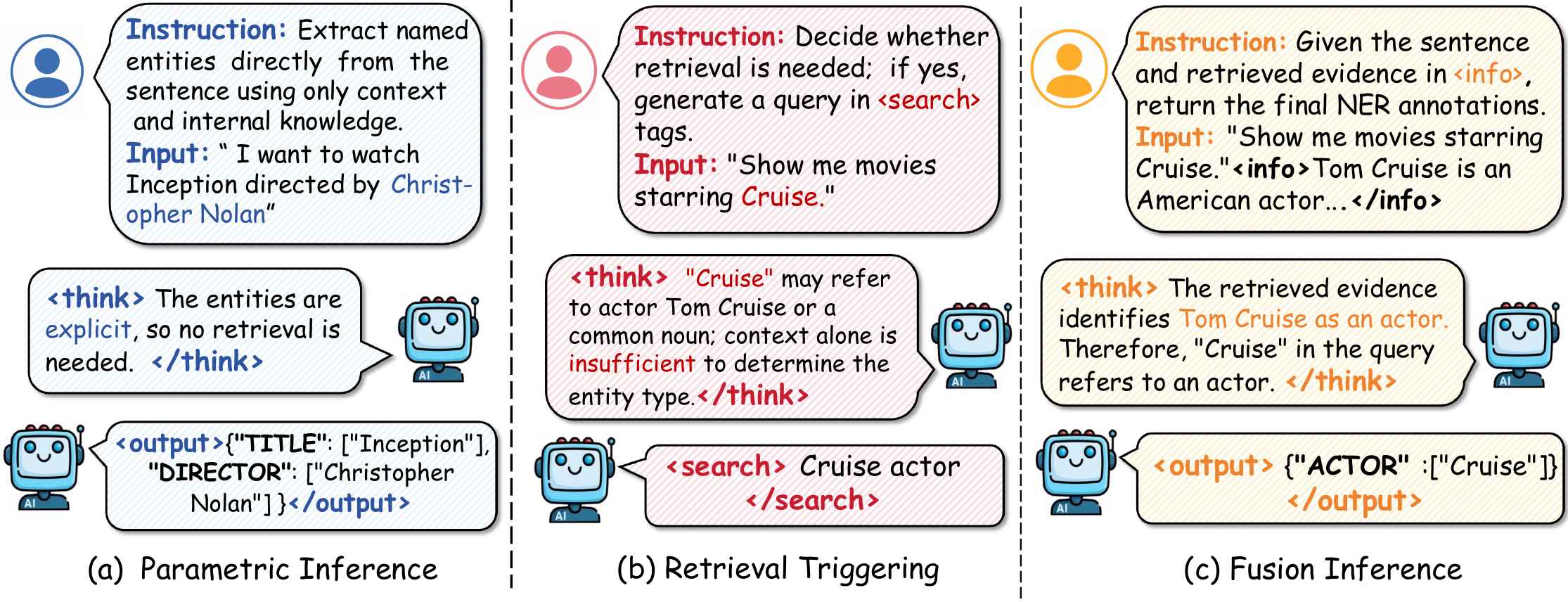}
    \caption{Illustration of the three supervised tasks in Multi-Task Capability Initialization (MTCI). (a) \textbf{Parametric Inference} performs NER directly from the input context and the model's parametric knowledge. (b) \textbf{Retrieval Triggering} determines whether external retrieval is necessary and, if needed, generates a query in \texttt{<search>} tags. (c) \textbf{Fusion Inference} integrates retrieved evidence to produce the final NER output while preserving spans from the original input.}
    \label{fig:mtci_tasks}
\end{figure*}

\paragraph{Task Design and Data Construction.}
We design three complementary instruction tuning tasks: parametric inference ($T_{\text{param}}$), retrieval triggering ($T_{\text{trigger}}$), and fusion inference ($T_{\text{rag}}$).

To construct high-quality training data that balances signal quality with sufficient exploration potential, we employ a pass-rate-based data selection mechanism. For each training sample, we use Qwen2.5-32B to generate $K{=}10$ candidate responses and apply task-specific selection criteria.

\noindent $\bullet$ \textbf{\textit{Parametric Inference}} ($T_{\text{param}}$) trains the model to extract entities using only the input context and its internal parametric knowledge. The target sequence contains an \texttt{<answer>} block with the predicted entities. For each training sample $x$, we estimate the non-retrieval pass rate as
\begin{equation}
    c_{\text{param}}(x) = \frac{1}{K}\sum_{k=1}^{K} 
    \mathbb{I}(\hat{y}_{\text{param},k} = y^*),
\end{equation}
where $\hat{y}_{\text{param},k}$ is the $k$-th sampled output, $y^*$ is the gold entity annotation, and $\mathbb{I}(\cdot)$ is the indicator function. Samples with reliable parametric predictions are retained:
\begin{equation}
    \mathcal{D}_{\text{param}} = \{x \mid c_{\text{param}}(x) 
    \geq \tau_{\text{param}}\}.
\end{equation}

\noindent $\bullet$ \textbf{\textit{Retrieval Triggering}} ($T_{\text{trigger}}$) trains the model to decide whether external knowledge is needed and, when necessary, to generate a rewritten search query. The supervised target contains a \texttt{<think>} fragment followed by a \texttt{<search>} block. The \texttt{<think>} fragment describes the source of uncertainty (e.g., entity ambiguity, rare mentions, or insufficient contextual evidence), while the \texttt{<search>} block contains an entity-centric rewritten query rather than the original sentence. Since no gold query exists for this task, we evaluate trigger quality indirectly via the downstream fusion pass rate (defined below) and retain trajectories whose retrieved evidence enables reliable fusion.

\noindent $\bullet$ \textbf{\textit{Fusion Inference}} ($T_{\text{rag}}$) trains the model to produce the final NER prediction after retrieved evidence is inserted via \texttt{<information>} tags. Given a query generated by $T_{\text{trigger}}$, the retriever returns evidence, and the teacher samples $K$ fusion outputs per training sample. We compute the retrieval-augmented pass rate as
\begin{equation}
    c_{\text{rag}}(x) = \frac{1}{K}\sum_{k=1}^{K} 
    \mathbb{I}(\hat{y}_{\text{rag},k} = y^*),
\end{equation}
where $\hat{y}_{\text{rag},k}$ is the $k$-th sampled fusion output. The fusion dataset is selected by
\begin{equation}
    \mathcal{D}_{\text{rag}} = \{x \mid c_{\text{rag}}(x) 
    \geq \tau_{\text{rag}}\}.
\end{equation}

Here, $\tau_{\text{param}}$, $\tau_{\text{rag}}$, and $\tau_{\text{trigger}}$ are task-specific selection thresholds. Beyond data filtering, the pass rates $c_{\text{param}}$ and $c_{\text{rag}}$ also partition training samples into easy and hard groups, which are used to define the retrieval-benefit reward in Section~\ref{sec:rlpo}.

\paragraph{Multi-Task Joint Fine-Tuning.} As illustrated in Figure~\ref{fig:mtci_tasks}, we perform multi-task joint fine-tuning on the filtered samples to acquire the three capabilities. We consolidate the tasks into $\mathcal{T} = \{T_{\text{param}}, T_{\text{trigger}}, T_{\text{rag}}\}$ with distinct supervision: $T_{\text{param}}$ generates answers using internal knowledge; $T_{\text{trigger}}$ generates retrieval queries; and $T_{\text{rag}}$ performs inference conditioned on retrieved information. The task-specific prompts are provided in Appendix~\ref{sec:appendix:mtci_prompts}.

We jointly optimize over this mixed distribution by minimizing the weighted negative log-likelihood loss:
\begin{equation}
\mathcal{L} = - \sum_{(x, y) \in \mathcal{T}} \lambda_{\text{type}} \sum_{t=1}^{|y|} \log P_\theta(y_t \mid y_{<t}, x),
\end{equation}
where $\lambda_{\text{type}}$ denotes the task-specific balancing weight, $x$ is the input context, $y$ is the corresponding target output sequence, and $|y|$ is its length.

\subsection{End-to-end RL Optimization}
\label{sec:rlpo}
MTCI equips the model with the abilities to reason, search, and fuse evidence, but it does not directly optimize the global trade-off between NER accuracy and retrieval cost. We therefore further optimize the initialized policy with rule-based rewards under a unified trajectory-level objective.

\paragraph{CoT-Guided Adaptive Retrieval.} To enable dynamic selection between parametric and retrieval-augmented reasoning, we design a Chain-of-Thought (CoT) guided paradigm that formalizes NER as a "reasoning-then-action" branching process. Given input $x$, the generation process is:
\begin{equation}
y = \begin{cases} 
\pi_\theta(y \mid x, c), & a(c) = \text{direct} \\[0.3em] 
\pi_\theta(y \mid x, c, \mathcal{R}(q)), & a(c) = \text{search}
\end{cases}
\end{equation}
where $c$ denotes the CoT reasoning fragment, $a(c)$ is the decision action, $q$ is the retrieval query, and $\mathcal{R}(q)$ denotes retrieved knowledge. This comprises two steps:

\noindent $\bullet$ \text{\textit{Introspective Planning.}} The model generates a \texttt{<think>} fragment $c$ to analyze entity ambiguity, domain knowledge needs, and context sufficiency. With insufficient internal knowledge, it generates a \texttt{<search>} tag with a rewrite search query $q$; otherwise, it proceeds directly to answer generation.

\noindent $\bullet$ \text{\textit{Evidence Fusion.}} When \texttt{<search>} is triggered, search engine $\mathcal{R}$ provides top-$k$ documents $\mathcal{R}(q)$. The model fuses input $x$ with $\mathcal{R}(q)$ then generates entity predictions $y$ in the \texttt{<answer>} block.
\paragraph{Multi-Dimensional Reward.}
We define a reward function that jointly encourages valid trajectories, accurate entity extraction, and selective retrieval.

\noindent $\bullet$ \textbf{\textit{Format Reward}} $r_{\text{fmt}}$ ensures that generated trajectories are parsable and verifiable:
\begin{equation}
 r_{\text{fmt}}(y) = (\omega_c + \omega_e)\, \mathbb{I}[\mathcal{F}(y)] - \omega_e,
\end{equation}
where $\mathcal{F}(y)$ checks whether tags such as \texttt{<search>} and \texttt{<answer>} are complete and properly paired, and $\omega_c, \omega_e > 0$ are the reward and penalty weights, respectively.

\noindent $\bullet$ \textbf{\textit{Accuracy Reward}} $r_{\text{acc}}$ uses the entity-level Micro-F1 between the prediction $y$ and the gold annotation $y^*$:
\begin{equation}
 r_{\text{acc}}(y, y^*) = \text{Micro-F1}(y, y^*).
\end{equation}

\noindent $\bullet$ \textbf{\textit{Retrieval Benefit Reward}} $r_{\text{bnf}}$ encourages retrieval on hard instances while discouraging unnecessary or harmful retrieval on easy instances:
\begin{equation}
\scalebox{0.86}{$\displaystyle
\begin{aligned}
 r_{\text{bnf}}(x, y) =
 \begin{cases}
 +w_{\text{hard}}^{\text{corr}}, & r_{\text{acc}} > \tau \land x \in \mathcal{X}_{\text{hard}}, \\
 +w_{\text{easy}}^{\text{corr}}, & r_{\text{acc}} > \tau \land x \in \mathcal{X}_{\text{easy}}, \\
 -w_{\text{hard}}^{\text{err}}, & r_{\text{acc}} \leq \tau \land x \in \mathcal{X}_{\text{hard}}, \\
 -w_{\text{easy}}^{\text{err}}, & r_{\text{acc}} \leq \tau \land x \in \mathcal{X}_{\text{easy}}.
 \end{cases}
\end{aligned}
$}
\end{equation}
We set $w_{\text{hard}}^{\text{corr}} > w_{\text{easy}}^{\text{corr}}$ and $w_{\text{hard}}^{\text{err}} < w_{\text{easy}}^{\text{err}}$, so that successful retrieval is rewarded more on hard instances while erroneous retrieval is penalized more on easy instances. The subsets $\mathcal{X}_{\text{hard}}$ and $\mathcal{X}_{\text{easy}}$ are determined from the MTCI pass rates. The final reward is
\begin{equation}
 R = \alpha\, r_{\text{acc}} + \gamma\, r_{\text{bnf}} + \lambda\, r_{\text{fmt}},
\end{equation}
where $\alpha$, $\gamma$, and $\lambda$ control the weights of accuracy, retrieval benefit, and format validity, respectively.

\paragraph{RL Algorithm.}
We optimize the policy using Group Relative Policy Optimization (GRPO). Unlike actor--critic methods, GRPO estimates advantages from a group of sampled trajectories and requires no separate critic network. Since retrieval evidence is externally injected rather than generated by the policy, we follow R1-Searcher++ \cite{song2025r1searcherplusplus} and mask environmental observation tokens when computing policy gradients:
\begin{equation}
 \mathbb{M}_t = \begin{cases}
 1, & y_t \in \mathcal{A}_{\text{gen}}, \\
 0, & y_t \in \mathcal{O}_{\text{env}},
 \end{cases}
\end{equation}
where $\mathcal{A}_{\text{gen}}$ denotes model-generated tokens and $\mathcal{O}_{\text{env}}$ denotes externally retrieved evidence. The masked GRPO objective is
\begin{equation}
 J(\theta) = \mathbb{E}\bigg[\frac{1}{G}\sum_{i=1}^{G} \mathcal{L}_i(\theta) - \beta D_{\text{KL}}\bigg],
\end{equation}
\begin{equation}
 \mathcal{L}_i(\theta) = \frac{1}{|y_i|}\sum_{t=1}^{|y_i|} \mathbb{M}_t \min(\rho_{i,t}\hat{A}_i,\, \bar{\rho}_{i,t}\hat{A}_i),
\end{equation}
where $\bar{\rho}_{i,t}=\text{clip}(\rho_{i,t},1-\epsilon,1+\epsilon)$, $G$ is the number of sampled trajectories, $\rho_{i,t}$ is the importance sampling ratio, $\hat{A}_i$ is the group-relative advantage, and $\beta$ is the KL penalty coefficient.

\section{Experiments}
In this section, we systematically evaluate the proposed NE-R1 framework to address the following research questions:

\begin{rQList}
    \item Does NE-R1 outperform existing models on NER benchmarks?
    \item Does NE-R1 generalize cross-domain in zero-shot?
    \item What is the contribution of each NE-R1 component?
    \item Does NE-R1 learn effective adaptive retrieval behavior?
\end{rQList}

\begin{table*}[!t]
\centering
\resizebox{1.0\textwidth}{!}{
\setlength{\tabcolsep}{3.5pt}
\renewcommand{\arraystretch}{1.08}
\begin{tabular}{@{}llccccc@{}}
\toprule
Category & Model & MIT-Movie & MIT-Rest. & OntoNotes & GENIA & Avg. \\
\midrule
\multirow{5}{*}{\makecell[l]{Few-shot\\LLMs}} & GPT-4o & 61.49 & 59.90 & 64.29 & 44.63 & 57.58 \\
 & Llama-3.3-70B & 60.76 & 53.62 & 59.98 & 41.91 & 54.07 \\
 & Qwen3-32B & 52.14 & 56.47 & 46.71 & 39.99 & 48.83 \\
 & DeepSeek-R1-MoE & 63.93 & 64.87 & 67.81 & 53.81 & 62.61 \\
\baserow & GPT-5 & 61.23 & 66.53 & 72.28 & 51.22 & 62.81 \\
\midrule
\multirow{5}{*}{\makecell[l]{NER\\baselines}} & InstructUIE~\cite{wang2023instructuie} & 89.58 & 82.59 & 88.64 & 75.71 & 84.13 \\
 & GLiNER-L~\cite{zaratiana2024gliner} & 87.90 & 83.60 & 89.80 & \underline{78.40} & 84.93 \\
 & $B^2$NER~\cite{yang2025beyond} & \underline{90.78} & \underline{83.71} & 84.31 & 76.43 & 83.81 \\
 & ReasoningNER~\cite{huang2026reasoning} & 89.60 & 81.40 & 89.10 & 78.30 & 84.60 \\
\baserow  & UniNER-7B~\cite{zhou2024universalner} & 90.17 & 82.35 & \underline{89.91} & 77.54 & \underline{85.00} \\
\midrule
\multirow{2}{*}{\makecell[l]{Retrieval\\baselines}} & Standard SFT (No Search) & 85.24 & 80.68 & 83.47 & 75.92 & 80.83 \\
\baserow & RAG (Always Search) & 85.56 & 81.72 & 84.38 & 77.23 & 82.22 \\
\midrule
\multirow{2}{*}{\makecell[l]{NE-R1\\variants}} & NE-R1 (MTCI only) & 87.94 & 83.46 & 85.12 & 77.77 & 83.57 \\
\dporow & NE-R1 (full) & \textbf{91.05}\dplus{+0.27} & \textbf{85.44}\dplus{+1.73} & \textbf{91.89}\dplus{+1.98} & \textbf{81.70}\dplus{+3.30} & \textbf{87.52}\dplus{+2.52} \\
\bottomrule
\end{tabular}
}
\caption{In-domain NER results. We report Micro-F1. Bold marks the best result in each column, underlining marks the strongest prior baseline, and the highlighted NE-R1 row additionally reports gains over the strongest prior result.}
\label{tab:main}
\end{table*}

\subsection{Setup}
\paragraph{Datasets and Metrics.}
We evaluate our model on standard benchmarks for both in-domain and cross-domain scenarios. For in-domain evaluation, we use OntoNotes~5.0 \cite{weischedel2013ontonotes}, MIT-Movie \cite{liu2013asgard}, MIT-Restaurant \cite{liu2013asgard}, and GENIA \cite{kim2003genia}, covering general, short-text, and specialized domains. For cross-domain evaluation (RQ2), following the standard protocol \cite{nandi2024improving}, models are trained on CoNLL-03 and evaluated zero-shot on the five CrossNER domains. All results are evaluated using strict entity-level Micro-F1. Detailed dataset statistics are provided in Appendix~\ref{sec:appendix:a1}.

\paragraph{Baselines.}
We benchmark NE-R1 against general-purpose LLMs (GPT-4o, GPT-5, Llama-3-70B, Qwen3-32B, DeepSeek-R1) and domain-specific NER models (InstructUIE, UniNER, GLiNER, $B^2$NER, ReasoningNER) for in-domain evaluation. For cross-domain evaluation, we follow~\cite{nandi2024improving} and adopt their baselines including transfer-based, prompting-based, and retrieval-augmented approaches.

\paragraph{Implementation Details.}
We use Qwen-2.5-7B as the backbone pre-trained language model, Qwen-2.5-32B generates training contents for MTCI. All experiments are conducted on 8$\times$NVIDIA A800 GPUs. In MTCI phase, we set learning rate to 1e-5 and batch size to 16. In RL phase, we set the rollout size to 5 and the learning rate to 1e-6. We employ a low-variance KL penalty (coefficient 1e-3) and a clip coefficient of 0.2. We use multi-source web search engines as external knowledge acquisition channels. For more detailed implementation details, refer to the appendix \ref{sec:appendix:a2}.

\begin{table*}[!t]
\centering
\resizebox{1.0\textwidth}{!}{
\setlength{\tabcolsep}{4pt}
\renewcommand{\arraystretch}{1.08}
\begin{tabular}{@{}lcccccc@{}}
\toprule
Model & Politics & Natural Science & Music & Literature & AI & Avg. \\
\midrule
BiLSTM-CRF~\cite{lample2016neural} & 56.60 & 49.97 & 44.79 & 43.03 & 43.56 & 47.59 \\
Coach~\cite{liu2020coach} & 61.50 & 52.09 & 51.66 & 48.35 & 45.15 & 51.75 \\
CROSS-DOMAIN LM~\cite{jia2019cross} & 68.44 & 64.31 & 63.56 & 59.59 & 53.70 & 61.92 \\
FLAIR~\cite{akbik2019flair} & 69.54 & 64.71 & 65.60 & 61.35 & 52.48 & 62.73 \\
BARTNER~\cite{liu2021crossner} & 69.90 & 65.14 & 65.35 & 58.93 & 53.00 & 62.46 \\
LIGHTNER~\cite{chen2022lightner} & 72.78 & 66.74 & 72.28 & 65.17 & 35.82 & 62.56 \\
LST-NER~\cite{zheng2022cross} & 73.25 & 70.07 & 76.83 & 70.76 & 63.28 & 70.84 \\
LANER~\cite{hu2022label} & 74.06 & 71.83 & 78.78 & 71.11 & 65.79 & 72.31 \\
\baserow CP-NER~\cite{chen2023one} & 74.25 & 75.82 & 79.10 & 72.17 & 67.95 & 73.86 \\
\midrule
GPT-NER~\cite{wang2025gpt} & 74.71 & 70.77 & 78.30 & 62.18 & 66.07 & 70.41 \\
PromptNER (GPT-3.5)~\cite{ashok2023promptner} & 71.74 & 64.83 & 77.78 & 64.15 & 59.35 & 67.57 \\
\baserow PromptNER (GPT-4)~\cite{ashok2023promptner} & 78.61 & 72.59 & 84.26 & 74.44 & 64.83 & 74.95 \\
\midrule
RAG + GPT-4 (sentence embeddings) & 78.20 & 73.52 & 83.61 & 71.32 & 66.91 & 74.71 \\
RAG + GPT-4 (word embeddings) & 78.63 & 73.95 & 84.25 & 74.68 & 68.19 & 75.94 \\
\baserow IF-WRANER~\cite{nandi2024improving} & \underline{79.80} & \underline{75.31} & \textbf{85.43} & \underline{75.52} & \underline{68.81} & \underline{76.97} \\
\midrule
\dporow NE-R1 (ours) & \textbf{81.35}\dplus{+1.55} & \textbf{75.83}\dplus{+0.52} & 83.51\dminus{-1.92} & \textbf{76.30}\dplus{+0.78} & \textbf{73.78}\dplus{+4.97} & \textbf{78.15}\dplus{+1.18} \\
\bottomrule
\end{tabular}
}
\caption{Zero-shot cross-domain NER results on CrossNER. We report Micro-F1. Bold marks the best result in each column, underlining marks the strongest prior baseline, and the highlighted NE-R1 row reports gains over IF-WRANER, the strongest overall baseline in average F1.}
\label{tab:cross_domain}
\end{table*}

\subsection{Main Results (In-Domain Evaluation)}
We first evaluate NE-R1 under the standard supervised setting. We compare the model against comparable (7B) and larger (70B+) baselines across four datasets characterized by diverse domain distributions and knowledge dependencies. As shown in Table~\ref{tab:main}, NE-R1 achieves superior performance on all benchmarks, consistently outperforming both equal-sized and larger models. Specifically, On MIT-Movie and MIT-Restaurant, it leads among 7B models with \textbf{91.05} and \textbf{85.44} F1 score, on the general-domain OntoNotes~5.0, it attains \textbf{91.89} F1 score, surpassing InstructUIE and UniNER-7B by \textbf{over 2.0 points}.  On the terminology-dense GENIA, it achieves \textbf{81.70} F1 score, a significant \textbf{3 point improvement} over GLiNER. These results demonstrate the effectiveness of adaptive retrieval
across the four evaluated in-domain NER benchmarks.

\begin{table}[!t]
\centering
\small
\setlength{\tabcolsep}{3.2pt}
\renewcommand{\arraystretch}{1.22}
\resizebox{\linewidth}{!}{%
\begin{tabular}{lccccc}
\toprule
Variant & Movie & Res. & Onto. & Gen. & Avg. \\
\midrule
\baserow Base model & 51.50 & 56.53 & 44.24 & 39.96 & 48.06 \\
w/o RL & 87.94 & 83.46 & 85.12 & 77.77 & 83.57 \\
w/o MTCI & 71.92 & 67.84 & 74.63 & 67.58 & 70.49 \\
w/o $T_{\text{trigger}}$\&$T_{\text{rag}}$ & 84.77 & 80.06 & 82.76 & 75.34 & 80.73 \\
w/o CoT & 89.84 & 84.73 & 89.37 & 80.71 & 86.16 \\
w/o $r_{\text{bnf}}$ & 88.71 & 83.68 & 86.64 & 79.39 & 84.61 \\
\addlinespace[1pt]
\dporow \textbf{NE-R1 (full)} & \textbf{91.05} & \textbf{85.44} & \textbf{91.89} & \textbf{81.70} & \textbf{87.52} \\
\bottomrule
\end{tabular}
}
\caption{Ablation study on different datasets. All results are micro-F1 score. Movie, Res., Onto., and Gen. denote MIT Movie, MIT Restaurant, OntoNotes, and GENIA datasets, respectively. ``w/o'' denotes removing a specific component from our full NE-R1 model.}
\label{tab:ablation}
\end{table}

\subsection{Cross-Domain Evaluation}

To address RQ2, we train NE-R1 on CoNLL-03 and evaluate it zero-shot on the five CrossNER domains (Table~\ref{tab:cross_domain}). Compared
with IF-WRANER, NE-R1 improves Politics, Natural Science, Literature, and AI by 1.55, 0.52, 0.78, and 4.97 F1 points, respectively, but performs 1.92 points lower on Music. It nevertheless achieves the best five-domain average of \textbf{78.15}, outperforming IF-WRANER by
\textbf{1.18} F1 points.

The largest gain on AI suggests that retrieval is particularly useful for specialized and evolving technical entities. In contrast, the smaller gains on Natural Science and Literature and the decrease on Music indicate that retrieval is less effective for errors involving
local context, entity boundaries, aliases, or dataset-specific label distinctions, and may introduce noise for ambiguous entity names. Overall, the benefit of adaptive retrieval varies across domains. 

\subsection{Ablation Studies}
To dissect the efficacy of NE-R1 (RQ3), we performed ablations on training stages, initialization tasks, and core components (Table~\ref{tab:ablation}).

\begin{figure*}[!t]
    \centering
    \begin{subfigure}[t]{0.58\linewidth}
        \centering
\includegraphics[width=\linewidth]{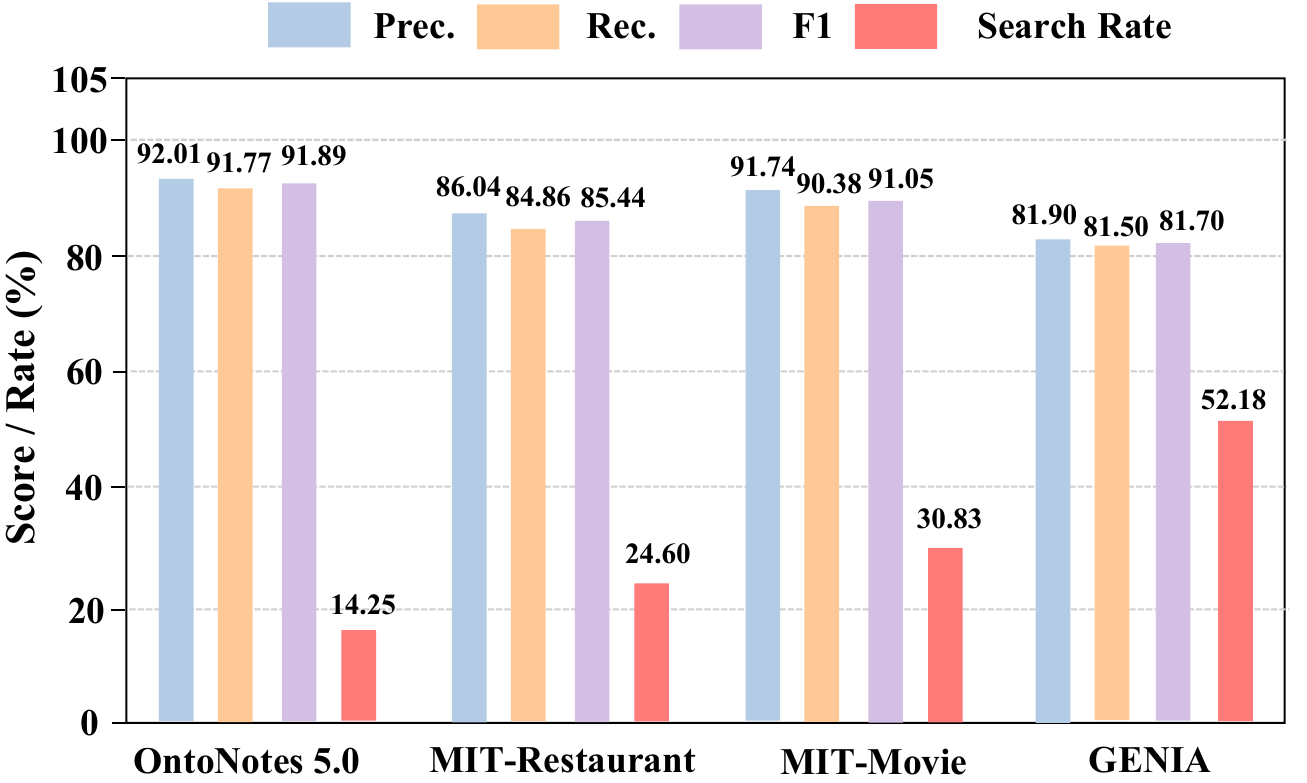}
        \caption{Domain-wise performance and search behavior of NE-R1.}
        \label{fig:domain_search}
    \end{subfigure}
    \hfill
    \begin{subfigure}[t]{0.40\linewidth}
        \centering
        \includegraphics[width=\linewidth]{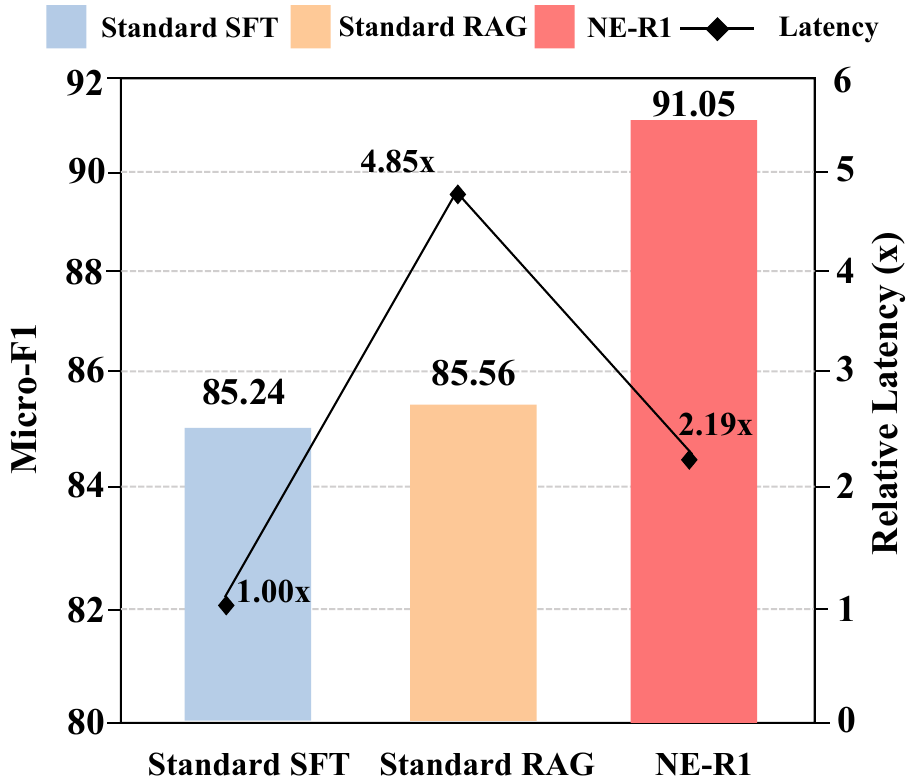}
        \caption{Efficiency comparison on MIT-Movie.}
        \label{fig:efficiency}
    \end{subfigure}
    \caption{Analysis of NE-R1's adaptive retrieval behavior. (a) NE-R1 calibrates retrieval frequency based on domain knowledge density. (b) NE-R1 achieves the best trade-off between accuracy and inference cost on MIT-Movie.}
    \label{fig:adaptive_analysis}
\end{figure*}

\begin{figure*}[!t] 
    \centering
    \begin{subfigure}[b]{0.48\textwidth} 
        \centering
        \includegraphics[width=\textwidth]{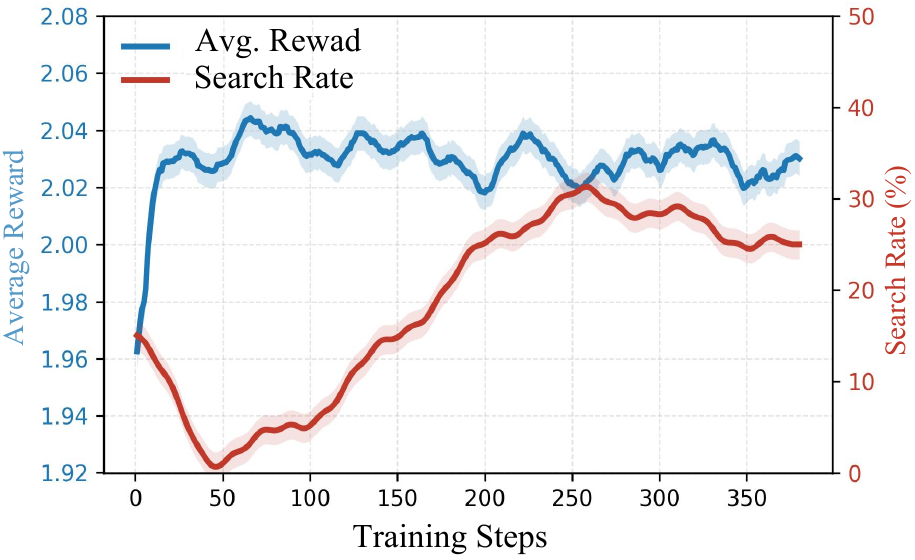} 
        \caption{Reward vs. Search Rate (Easy Samples)}
        \label{fig:easy_samples}
    \end{subfigure}
    \hfill 
    \begin{subfigure}[b]{0.48\textwidth}
        \centering
        \includegraphics[width=\textwidth]{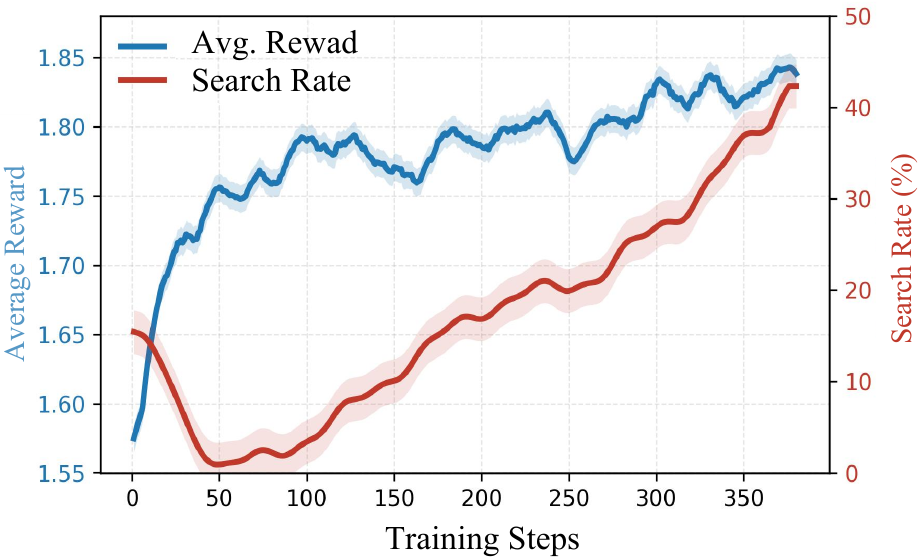}
        \caption{Reward vs. Search Rate (Hard Samples)}
        \label{fig:hard_samples}
    \end{subfigure}
    
    \caption{Reward and search rate dynamics under different sample difficulties. 
    Note that the blue line represents Average Reward and the red line represents Search Rate. }
    \label{fig:reward_search}
\end{figure*}

\paragraph{Training Stage Ablation.} \textit{w/o MTCI} removes the Multi-Task Capability Initialization stage; \textit{w/o end-to-end RL} removes the RL Optimization stage. Removing MTCI prevents effective query generation and knowledge integration, while removing end-to-end RL causes notable performance drops, confirming both stages are essential.

\paragraph{Initialization Task Ablation.} \textit{w/o $T_{\text{trigger}}$ and $T_{\text{rag}}$} excludes Retrieval Triggering and Fusion Inference from initialization tasks. This causes a 6.8 point drop in F1 score (87.52$\rightarrow$80.73), demonstrating that joint training on query generation and information fusion is crucial.

\paragraph{Reasoning and Reward Ablation.} \textit{w/o CoT} removes Chain-of-Thought reasoning; \textit{w/o $r_{\text{bnf}}$} removes Retrieval Benefit Reward. Ablating $r_{\text{bnf}}$ proves more detrimental than removing CoT, indicating the benefit-aware reward is central to balancing retrieval costs.

\subsection{Analysis of Adaptive Search Behavior}

To address RQ4, we analyze the retrieval behavior of NE-R1 from three
perspectives: domain-level retrieval frequency, the efficiency--performance
trade-off, and training dynamics. Together, these analyses examine
whether retrieval decisions vary with knowledge demand rather than
following a fixed always-retrieve or never-retrieve policy.

\paragraph{Search Behavior Divergence.}
Figure~\ref{fig:adaptive_analysis}(a) compares retrieval rates across
domains with different knowledge requirements. The retrieval rate is
14.25\% on OntoNotes 5.0, where many entities can be recognized from
local context and parametric knowledge. It increases to 24.60\% on
MIT-Restaurant and 30.83\% on MIT-Movie, whose short queries often
contain ambiguous names with limited contextual information. On GENIA,
the rate reaches 52.18\%, reflecting the greater need for external
knowledge when recognizing specialized biomedical terminology. This
variation suggests that NE-R1 adjusts its retrieval frequency to the
knowledge demands of different domains instead of applying a uniform
retrieval policy.

\paragraph{Speed--Quality Trade-off.}
We compare NE-R1 with Standard SFT and Standard RAG (always-retrieve)
on MIT-Movie, as shown in Figure~\ref{fig:adaptive_analysis}(b).
Standard SFT achieves 85.24 Micro-F1 with a relative latency of
1.00$\times$, whereas Standard RAG achieves only a modest improvement
to 85.56 Micro-F1 while increasing latency to 4.85$\times$. In
comparison, NE-R1 achieves the highest Micro-F1 score of 91.05 with a
relative latency of 2.19$\times$, corresponding to an approximately
55\% reduction in inference latency relative to Standard RAG. These
results show that adaptive retrieval avoids part of the overhead of
indiscriminate retrieval while obtaining more useful external evidence.

\paragraph{Reward Evolution.}
Figure~\ref{fig:reward_search} illustrates the evolution of average
reward and retrieval rate for easy and hard samples. For easy samples
(Figure~\ref{fig:easy_samples}), the retrieval rate decreases while
the reward remains high, indicating that the model increasingly relies
on parametric knowledge and avoids unnecessary retrieval. For hard
samples (Figure~\ref{fig:hard_samples}), the retrieval rate increases
as training progresses, allocating more retrieval to inputs for which
external evidence is more likely to be useful. The opposing trends
show that training does not simply encourage a globally higher or
lower retrieval rate. Instead, they support that NE-R1 learns a
difficulty-aware retrieval strategy conditioned on the expected
utility of external evidence.


\subsection{Retrieval Leakage Audit}

NE-R1 does not directly retrieve test answers or benchmark annotations. The model generates entity-centric queries around uncertain mentions only when external knowledge is needed, rather than submitting complete test sentences to search engines. The retrieved
content provides entity-related background knowledge for semantic disambiguation, while predicted entity spans and types remain constrained by the original input and the predefined label set.

We conduct an exhaustive leakage audit across all four in-domain benchmarks: MIT-Movie, MIT-Restaurant, OntoNotes 5.0, and GENIA. Their
test sets contain 1,953, 1,521, 8,262, and 1,854 instances, respectively, covering 13,590 test instances in total. For every instance that triggers retrieval, we manually inspect the generated query and all of its top-3 retrieved snippets. The exhaustive audit
covers \textbf{3,120} retrieval-triggered instances and \textbf{9,360} query--snippet pairs in total.

We examine whether the retrieved content contains annotation-related terms, BIO tags or other structured NER outputs, gold entity lists, complete test sentences, or links to benchmark pages and annotation files. We find no cases in which retrieval causes test-answer leakage.

\section{Conclusion}

In this paper, we introduced NE-R1, an adaptive retrieval-augmented framework for NER that implements a \textbf{``retrieval-on-demand''} mechanism. NE-R1 first acquires parametric inference, retrieval
triggering, and evidence-fusion capabilities through multi-task instruction tuning, and then jointly optimizes NER predictions and retrieval decisions through end-to-end reinforcement learning. This design allows the model to rely on parametric knowledge for familiar
inputs and retrieve external evidence for knowledge-intensive or ambiguous entities.

Across the four evaluated in-domain benchmarks, NE-R1 improves the average F1 score by 2.52 points over the strongest overall baseline. In zero-shot transfer from CoNLL-03 to CrossNER, it improves performance in four of the five domains and achieves the best five-domain average, exceeding the strongest baseline by 1.18 F1 points. Behavioral analysis further
shows that retrieval varies across domains and sample difficulties, while on MIT-Movie, NE-R1 reduces inference latency by approximately 55\% relative to always-retrieve RAG. Overall, these results demonstrate that NE-R1 consistently outperforms existing generative and retrieval-augmented NER methods with higher efficiency, while generalizing robustly to unseen entity types.

\section*{Limitations}

First, NE-R1 relies on live, multi-source Web search. The retrieved content and its ranking may change over time because of API updates,
indexing changes, and regional variation. Consequently, exact reproduction with live search services cannot be guaranteed, even
under the same retrieval configuration. 

Second, our empirical conclusions are limited to the evaluated English NER benchmarks and the current multi-source Web-search setting. The experiments do not establish that NE-R1 generalizes to other languages, domains, or retrieval backends. Moreover, heterogeneous search engines may return multilingual, conflicting, or uneven-quality evidence, and
we do not separately quantify the effects of these factors.

Finally, the current framework supports only single-turn retrieval. While iterative retrieval may provide additional evidence for obscure or highly ambiguous entities, it may also introduce additional latency, noise, and error propagation. The main results follow the standard strict entity-level Micro-F1 evaluation protocol. Because repeated-run
results or test predictions are unavailable for many comparison systems, we cannot conduct a uniform run-level variance or paired
significance analysis across all baselines.

\bibliography{custom_v2}

\clearpage
\appendix

\renewcommand{\dbltopfraction}{0.85}
\renewcommand{\dblfloatpagefraction}{0.95}
\renewcommand{\textfraction}{0.10}

\section{Appendix}
\label{sec:appendix}

\subsection{Dataset Statistics}
\label{sec:appendix:a1}

Table \ref{tab:dataset_stats} presents the detailed statistics of the datasets used in our experiments, including the number of sentences in train/dev/test splits, the number of entity types, the average token length, and the average number of entities per sentence.

\begin{table*}[!tbp]
\centering
\small
\renewcommand{\arraystretch}{1.15}
\resizebox{\dimexpr2\columnwidth+\columnsep\relax}{!}{%
\begin{tabular}{lcccccc}
\toprule
\textbf{Dataset} & \textbf{\# train} & \textbf{\# dev} & \textbf{\# test} & \textbf{\# types} & \textbf{Avg. tokens} & \textbf{Avg. entities} \\
\midrule
Ontonotes\cite{weischedel2013ontonotes} & 59924 & 8528 & 8262 & 18 & 18 & 0.9 \\
GENIA\cite{kim2003genia} & 15023 & 1669 & 1854 & 5 & 43 & 3.5 \\
MIT-Movie\cite{liu2013asgard} & 6816 & 1000 & 1953 & 12 & 11 & 2.0 \\
MIT-Restaurant\cite{liu2013asgard} & 7660 & 1000 & 1521 & 8 & 9 & 1.7 \\
CoNLL-03\cite{sang2003introduction} & 14041 & 3250 & 3453 & 3 & 25 & 1.9 \\
\midrule
CrossNER AI\cite{liu2021crossner} & 100 & 350 & 431 & 13 & 52 & 5.3 \\
CrossNER Literature\cite{liu2021crossner} & 100 & 400 & 416 & 11 & 54 & 5.4 \\
CrossNER Music\cite{liu2021crossner} & 100 & 380 & 465 & 12 & 57 & 6.5 \\
CrossNER Politics\cite{liu2021crossner} & 199 & 540 & 650 & 8 & 61 & 6.5 \\
CrossNER Science\cite{liu2021crossner} & 200 & 450 & 543 & 16 & 54 & 5.4 \\
\bottomrule
\end{tabular}%
}
\caption{Dataset statistics.}
\label{tab:dataset_stats}
\end{table*}

\paragraph{In-Domain Datasets.} We utilize a diverse set of benchmarks to assess generalizability across different data distributions:

\begin{itemize}
    \item \textbf{Standard \& General Domain:} \textbf{CoNLL-03} \cite{sang2003introduction} and \textbf{Ontonotes 5.0} \cite{weischedel2013ontonotes} represent standard news and general text domains. Ontonotes 5.0 is the largest dataset in our evaluation, containing over 59k training samples with 18 fine-grained entity types.
    
    \item \textbf{Specialized Domain:} \textbf{GENIA} \cite{kim2003genia} focuses on the biomedical domain. As shown in Table \ref{tab:dataset_stats}, it exhibits a high density of entities (Avg. entities = 3.5), reflecting the knowledge-intensive nature of scientific texts.
    
    \item \textbf{Short-Text Domain:} \textbf{MIT-Movie} and \textbf{MIT-Restaurant} \cite{liu2013asgard} consist of informal user queries. These datasets have significantly shorter sentence lengths (Avg. tokens $\approx$ 9--11) compared to standard news corpora, posing challenges for context-dependent entity recognition.
\end{itemize}

\paragraph{Cross-Domain Datasets.} For the cross-domain evaluation scenarios (trained on CoNLL-03), we employ the \textbf{CrossNER} benchmark \cite{liu2021crossner}. This dataset encompasses five distinct domains: \textit{AI, Literature, Music, Politics,} and \textit{Science}. While our main experiment primarily follows a zero-shot setting, Table \ref{tab:dataset_stats} lists the standard data splits provided by the benchmark, which are typically used for low-resource or few-shot adaptation studies.

\subsection{Implementation Details}
\label{sec:appendix:a2}

\paragraph{Hyperparameters.} In Table \ref{tab:hyperparams}, we present detailed values of hyperparameters used in our study, categorized into the Multi-Task Capability Initialization (MTCI) phase and the Reinforcement Learning (RL) phase. All experiments were conducted on 8 NVIDIA A800 GPUs using Qwen-2.5-7B as the backbone model.

\begin{table}[!t]
\centering
\small
\begin{tabular}{lc}
\hline
\textbf{Hyperparameter} & \textbf{Value} \\
\hline
\multicolumn{2}{c}{\textit{MTCI Phase}} \\
Learning Rate & 1e-5 \\
LR Scheduler & Cosine \\
Warmup Ratio & 0.1 \\
Num Epochs & 5 \\
Per-Device Batch Size & 16 \\
Gradient Accumulation & 1 \\
\hline
\multicolumn{2}{c}{\textit{RL Phase }} \\
Optimizer & AdamW \\
Learning Rate & 1e-6 \\
Warmup Ratio & 0.01 \\
Total Steps & 500 \\
Data Collection Batch Size & 512 \\
Clip Ratio & 0.2 \\
KL Coefficient & 0.001 \\
KL Type & low\_var\_kl \\
\textbf{Rollout \& Generation} & \\
Max Prompt Length & 4096 \\
Max Response Length & 500 \\
Max Turns & 2 \\
Temperature & 0.95 \\
Rollout Samples ($N$) & 5 \\
Retriever Top-$k$ & 3 \\
\hline
\end{tabular}
\caption{Detailed hyperparameter configuration for both MTCI and RL phases.}
\label{tab:hyperparams}
\end{table}

In the SFT phase, we utilized the AdamW optimizer with a learning rate of 1e-5 and a cosine learning rate scheduler. The training was conducted for 5 epochs with a warmup ratio of 0.1. We set the per-device batch size to 16 with gradient accumulation steps set to 1, and enabled bfloat16 precision to optimize memory usage.

In the RL phase, the model was further optimized using Group Relative Policy Optimization (GRPO), without introducing a critic network. We lowered the learning rate to 1e-6 with a constant scheduler and a 1\% warmup phase. The training spanned 500 steps with a data collection batch size of 512, and GRPO policy updates were performed on mini-batches of size 256. The clipping ratio was set to 0.2, and we incorporated a KL divergence penalty with a coefficient of 0.001, utilizing a low-variance estimator (low\_var\_kl). Regarding model configuration, we limited the maximum prompt length to 4096 tokens and the response length to 500 tokens, supporting up to 2 interaction turns. During the rollout phase, we set the sampling temperature to 0.95 and generated 5 candidate responses per prompt (N=5). For retrieval-augmented generation, the top-3 relevant documents were retrieved. Some implementation frameworks retain generic policy-optimization parameter names such as mini-batch or clip-ratio settings; in our experiments, the actual advantage estimator is GRPO.

\paragraph{Search Engines.} In the implementation of the retrieval module, we utilize multi-source web search engines (as shown in the table \ref{tab:search_engines}) as the external knowledge acquisition channels, and dynamically incorporate the retrieval results into the reasoning process of the model through a unified search interface.

\begin{table}[!t]
\centering
\small
\begin{tabular}{lcp{5.2cm}}
\toprule
\textbf{Search Engine} & \textbf{URL} \\
\midrule
Baidu Search  & \url{https://www.baidu.com} \\
Google Search  & \url{https://www.google.com} \\
Bing Search   & \url{https://www.bing.com} \\
Wikipedia    & \url{https://www.wikipedia.org} \\
\bottomrule
\end{tabular}
\caption{Search engines used in the retrieval module. All search engines are accessed through a unified retrieval interface.}
\label{tab:search_engines}
\end{table}

\subsection{Supplementary Experiments and Analyses}
\label{sec:appendix:supp_analysis}

\paragraph{Matched-rate random trigger baseline.}
To further isolate the effect of the learned retrieval policy, we compare NE-R1 with random retrieval policies under different fixed search rates. As shown in Table~\ref{tab:random_trigger_appendix}, random triggering improves over naive retrieval decisions in some settings, but remains consistently inferior to the adaptive policy of NE-R1. This confirms that NE-R1 does not merely benefit from using a certain retrieval budget; instead, its learned trigger policy better identifies when external evidence is needed.

\begin{table*}[!tbp]
\centering
\small
\setlength{\tabcolsep}{5pt}
\renewcommand{\arraystretch}{1.15}
\begin{tabular}{lcccccc}
\toprule
\textbf{Method} & \textbf{Search Rate} & \textbf{MIT-Movie} & \textbf{MIT-Rest.} & \textbf{OntoNotes} & \textbf{GENIA} & \textbf{Avg.} \\
\midrule
NE-R1 + Random & 20\% & 88.73 & 84.51 & 88.76 & 78.94 & 85.24 \\
NE-R1 + Random & 35\% & 89.61 & 84.19 & 87.83 & 79.87 & 85.38 \\
NE-R1 + Random & 50\% & 88.12 & 83.44 & 86.71 & 80.63 & 84.73 \\
\textbf{NE-R1 (Adaptive)} & 14--52\% & \textbf{91.05} & \textbf{85.44} & \textbf{91.89} & \textbf{81.70} & \textbf{87.52} \\
\bottomrule
\end{tabular}
\caption{Comparison between random retrieval triggering and NE-R1's adaptive retrieval policy. Random triggering uses fixed retrieval rates, while NE-R1 adaptively adjusts its search rate across datasets according to input difficulty and domain knowledge requirements. All results are Micro-F1 scores.}
\label{tab:random_trigger_appendix}
\end{table*}

\paragraph{Discussion on retrieval leakage risk.}
A potential concern for retrieval-augmented evaluation on public NER benchmarks is that a search engine may return benchmark annotations or otherwise create an ``open-book'' setting. NE-R1 mitigates this risk in three ways. First, the emitted query is not the original test sentence; it is an entity-centric rewritten query produced inside the \texttt{<search>} tag, such as a request for background information about an ambiguous mention. Second, the retrieved content is inserted as natural-language evidence through \texttt{<information>} tags and does not provide BIO labels or dataset-specific NER annotations. Third, the always-retrieve baseline does not dominate NE-R1, suggesting that merely increasing retrieval exposure is insufficient; the benefit comes from deciding when the external evidence is useful. The zero-shot cross-domain evaluation further supports that the model relies on transferable retrieval and reasoning behavior rather than memorizing benchmark-specific labels.

\subsection{MIT-Movie Prompt Templates}
\label{sec:appendix:mtci_prompts}
This section presents the prompt templates used in our MTCI and evaluation pipelines, using MIT-Movie as a representative example. The same prompting scheme is applied to other datasets by adapting the task description and entity label set accordingly. Specifically, the task prompts for parametric inference, retrieval triggering, and fusion inference, together with the evaluation-time adaptive retrieval prompt, are shown in figures~\ref{fig:mitmovie_task1_prompt},
  \ref{fig:mitmovie_task2_prompt}, \ref{fig:mitmovie_task3_prompt}, and \ref{fig:mitmovie_eval_prompt}, respectively.

\paragraph{Parametric inference prompt.} This prompt initializes direct NER extraction without retrieval. The examples emphasize preserving the original span text and returning a compact JSON annotation.

\begin{figure*}[!tbp]
\centering
\begin{tcolorbox}[
    enhanced,
    title={\textbf{MIT-Movie Task 1: Parametric Inference ($T_{\text{param}}$)}},
    colframe=headerBlue,
    colback=bgBlue,
    colbacktitle=headerBlue,
    coltitle=white,
    fonttitle=\bfseries,
    arc=2mm,
    boxrule=0.7pt,
    width=\textwidth,
    left=2mm,right=2mm,top=1.5mm,bottom=1.5mm]
\scriptsize
\sffamily
You are a senior NER expert. Please extract entity spans from the given query.\\[0.2em]
\textbf{Entity Categories:} ACTOR: real-world actors' names; CHARACTER: movie characters; DIRECTOR: film directors' names; GENRE: movie genres, such as action, comedy, horror, mystery, and 3D; PLOT: storyline phrases, such as vampires, ghosts, war, or explosions; RATING: age-rating labels; RATINGS\_AVERAGE: scores, star ratings, or recommendation levels; REVIEW: subjective review terms; SONG: song titles; TITLE: official movie titles; TRAILER: trailer or preview mentions; YEAR: years or time expressions related to movies.\\[0.2em]
\textbf{Output Rules:} (1) Extract entity spans according to context and keep the original surface text. (2) Return a JSON string whose keys are entity categories and whose values are span lists. (3) Omit empty categories.\\[0.2em]
\textbf{Examples:}\\
Query: on what film did brandon lee last seen\\
Output: \{"ACTOR": ["brandon lee"]\}\\
Query: where did george clooney film the ides of march\\
Output: \{"ACTOR": ["george clooney"], "TITLE": ["the ides of march"]\}\\
Query: is there a mystery movie featuring a f silver\\
Output: \{"GENRE": ["mystery"], "DIRECTOR": ["a f silver"]\}\\
Query: list the sandlot movies\\
Output: \{"TITLE": ["sandlot"]\}\\[0.2em]
\textbf{Start:} Query: what movie has the shortest trailer\\
Output:
\end{tcolorbox}
\caption{MIT-Movie prompt for the parametric inference task. This task trains the model to perform NER directly without external retrieval.}
\label{fig:mitmovie_task1_prompt}
\end{figure*}

\paragraph{Retrieval triggering prompt.} This prompt teaches the model to separate clear cases from ambiguous or knowledge-intensive ones, producing a search query only when external evidence is likely to help.

\begin{figure*}[!tbp]
\centering
\begin{tcolorbox}[
    enhanced,
    title={\textbf{MIT-Movie Task 2: Retrieval Triggering ($T_{\text{trigger}}$)}},
    colframe=headerGreen,
    colback=bgGreen,
    colbacktitle=headerGreen,
    coltitle=white,
    fonttitle=\bfseries,
    arc=2mm,
    boxrule=0.7pt,
    width=\textwidth,
    left=2mm,right=2mm,top=1.5mm,bottom=1.5mm]
\scriptsize
\sffamily
You are a senior NER expert. Determine whether search is necessary before MIT-Movie NER.\\[0.2em]
\textbf{Search Rules:} (1) Search when the context cannot determine whether a span belongs to an entity category. (2) Search when a mention is ambiguous, e.g., it may refer to an actor, director, character, common noun, or movie title. (3) Do not search when the intent and entity type are clear from context.\\[0.2em]
\textbf{Examples:}\\
Query: matrix\\
Thought: ``matrix'' may be a movie title; search can verify it.\\
Search Query: matrix movie\\
Query: 3d\\
Thought: In the movie domain, ``3d'' clearly refers to a genre or format; no search is needed.\\
Search Query: \\
Query: what is brave about\\
Thought: ``Brave'' may be a movie title; search can verify it.\\
Search Query: Brave movie\\
Query: did ray liotta direct any films\\
Thought: The mention is clearly a person name; no search is needed for NER.\\
Search Query: \\[0.2em]
\textbf{Output Format:} $<$think$>$...$<$/think$>$; $<$thought$>$...$<$/thought$>$; $<$search\_q$>$...$<$/search\_q$>$.\\[0.2em]
\textbf{Start:} Query: what movies star bruce willis\\
Thought:
\end{tcolorbox}
\caption{MIT-Movie prompt for the retrieval triggering task. This task initializes the model to reason about retrieval necessity and generate search queries.}
\label{fig:mitmovie_task2_prompt}
\end{figure*}

\paragraph{Fusion inference prompt.} This prompt trains evidence-aware extraction. It explicitly instructs the model to use retrieved snippets for disambiguation while keeping final entity spans grounded in the original query.

\begin{figure*}[!tbp]
\centering
\begin{tcolorbox}[
    enhanced,
    title={\textbf{MIT-Movie Task 3: Fusion Inference ($T_{\text{rag}}$)}},
    colframe=headerBlue,
    colback=DataYellow,
    colbacktitle=headerBlue,
    coltitle=white,
    fonttitle=\bfseries,
    arc=2mm,
    boxrule=0.7pt,
    width=\textwidth,
    left=2mm,right=2mm,top=1.5mm,bottom=1.5mm]
\scriptsize
\sffamily
You are a senior NER expert. Please extract entity spans from the given MIT-Movie query. Search results are provided as external evidence; use them to resolve ambiguous names, long-tail movie titles, actors, directors, characters, and other knowledge-intensive mentions. Do not copy irrelevant evidence into the answer.\\[0.2em]
\textbf{Entity Categories:} ACTOR: real-world actors' names; CHARACTER: movie characters; DIRECTOR: film directors' names; GENRE: movie genres; PLOT: storyline phrases; RATING: age-rating labels; RATINGS\_AVERAGE: scores, star ratings, or recommendation levels; REVIEW: subjective review terms; SONG: song titles; TITLE: official movie titles; TRAILER: trailer or preview mentions; YEAR: years or time expressions related to movies.\\[0.2em]
\textbf{Output Rules:} (1) Extract spans from the original query and keep their original surface form. (2) Return a JSON string whose keys are entity categories and whose values are span lists. (3) Omit empty categories. (4) Use retrieved evidence only to determine the entity type; the final span must come from the query rather than the retrieved snippets.\\[0.2em]
\textbf{Retrieved Evidence:} WebSearch(key="Harold Ramis actor Bill Murray actor"): 1. Harold Ramis is a director, actor, and writer. Representative works include Groundhog Day. 2. Bill Murray is an actor and comedian. Representative works include Groundhog Day and Ghostbusters. 3. Groundhog Day cast information lists Harold Ramis and Bill Murray.\\[0.2em]
\textbf{Start Annotating:} Query: find me all of the movies that starred harold ramis and bill murray\\
Output: \{"ACTOR": ["harold ramis", "bill murray"]\}
\end{tcolorbox}
\caption{MIT-Movie prompt for the fusion inference task. Retrieved evidence is provided to help the model resolve ambiguous or knowledge-intensive mentions before producing the final NER output.}
\label{fig:mitmovie_task3_prompt}
\end{figure*}

\paragraph{Evaluation prompt.} The evaluation prompt unifies the direct and retrieval-augmented paths. The model may search when necessary, but the final answer must always be a valid JSON annotation inside the answer tag.

\begin{figure*}[!tbp]
\centering
\begin{tcolorbox}[
    enhanced,
    title={\textbf{MIT-Movie Evaluation Prompt with Adaptive Retrieval}},
    colframe=seedblue,
    colback=bgBlue,
    colbacktitle=seedblue,
    coltitle=white,
    fonttitle=\bfseries,
    arc=2mm,
    boxrule=0.7pt,
    width=\textwidth,
    left=2mm,right=2mm,top=1.5mm,bottom=1.5mm]
\scriptsize
\sffamily
You are a senior NER expert. Please extract entity spans from the given MIT-Movie query. You may either answer directly or call a search engine when external knowledge is needed.\\[0.2em]
\textbf{Entity Categories:} ACTOR: real-world actors' names; CHARACTER: characters in movies; DIRECTOR: film directors' names; GENRE: movie genres, such as action, comedy, horror, and 3D; PLOT: phrases describing the storyline, such as vampires, ghosts, love at first sight, war, or explosions; RATING: age-rating labels, e.g., pg 13 or nc 17; RATINGS\_AVERAGE: scores, star ratings, or recommendation levels; REVIEW: subjective review terms; SONG: song titles; TITLE: official movie titles; TRAILER: trailer or preview mentions; YEAR: years or time expressions related to movies.\\[0.2em]
\textbf{Extraction Rules:} (1) Extract spans based on query context and keep the original text. (2) Return a JSON string whose keys are entity categories and whose values are span lists. (3) Omit empty categories. (4) If search is used, use retrieved information only as evidence for disambiguation; do not output entities that do not appear in the query.\\[0.2em]
\textbf{Search Rules:} Call search when the context cannot determine whether a span belongs to one of the entity categories, or when a mention is ambiguous, such as a name that may refer to an actor, director, character, common noun, or movie title. Do not search if the query semantics are already clear.\\[0.2em]
\textbf{Output Format:} Need to search: $<$search$>$search query$<$/search$>$ $<$information$>$search info...$<$/information$>$ $<$answer$>$final answer$<$/answer$>$. No need to search: $<$answer$>$final answer$<$/answer$>$. The final answer must be valid JSON inside the answer tag.\\[0.2em]
\textbf{Start Annotating:} Query: \{user\_input\}
\end{tcolorbox}
\caption{MIT-Movie evaluation prompt with adaptive retrieval. The model either triggers search and fuses retrieved information, or directly returns the final NER answer.}
\label{fig:mitmovie_eval_prompt}
\end{figure*}

\subsection{Case Study}
\label{sec:appendix:a3}
\paragraph{Successful adaptive retrieval behavior.}
Figure~\ref{fig:casestudy} visualizes the inference paths of NE-R1 across varying sample difficulties, demonstrating its capability for uncertainty-driven on-demand retrieval. In the ambiguous Case 1, the span "Crash" is polysemous (common noun vs. movie title). Recognizing that context alone is insufficient, the model flags uncertainty and triggers retrieval to resolve the ambiguity, correctly identifying it as a Title. Conversely, in the confident Case 2 ("Steven Spielberg"), the model detects a clear intent pattern and performs direct prediction without retrieval overhead. This comparison indicates that retrieval decisions are driven by introspective judgment rather than heuristics: the model seeks external knowledge only when internal evidence is lacking, bypassing retrieval for simple cases to balance efficiency and accuracy. This validates that NE-R1 has effectively learned an intuitive "retrieval-on-demand" strategy. The corresponding visualized trajectories are shown in Figure~\ref{fig:casestudy}.

\begin{figure*}[!tbp]
    \centering
    \small
    \sffamily

    \begin{tcolorbox}[
        enhanced,
        title={\textbf{Case 1: Ambiguity Resolution (Hard Sample)}}, 
        colframe=headerBlue,     
        colbacktitle=headerBlue, 
        coltitle=white,          
        colback=bgBlue,          
        fonttitle=\bfseries,     
        arc=2mm,                 
        boxrule=0.8pt,           
        left=2.5mm, right=2.5mm, top=2.5mm, bottom=2.5mm, 
        boxsep=1.2mm
    ]
        \textbf{Input :} ``\textit{I am looking for information about \underline{Crash}.}''
        
        \textbf{Model Inference:}
        \begin{itemize}[leftmargin=1.5em, label=\textbullet, nosep] 
            \item $<$think$>$ : The span ``Crash'' is ambiguous. It can refer to a common noun (accident) or a movie title... the type (TITLE vs non-entity) is unclear. Context alone is insufficient.$<$/think$>$ \textbf{\textcolor{decisionRed}{[Decision:SEARCH]}}
            \item $<$search$>$Crash movie information$<$/search$>$
            \item $<$information$>$Search results show that ``Crash'' is a well-known movie title, including: Crash (1996), directed by David Cronenberg
        \end{itemize}
        
        \textbf{Output :} $<$answer$>$(Crash, TITLE)$<$/answer$>$
    \end{tcolorbox}

    \begin{tcolorbox}[
        enhanced,
        title={\textbf{Case 2 : Confident Prediction (Easy Sample)}}, 
        colframe=headerGreen,
        colbacktitle=headerGreen,
        coltitle=white,
        colback=bgGreen,
        fonttitle=\bfseries,
        arc=2mm,
        boxrule=0.8pt,
        left=2.5mm, right=2.5mm, top=2.5mm, bottom=2.5mm,
        boxsep=1.2mm
    ]
        \textbf{Input :} ``\textit{find movies directed by \underline{steven spielberg}}''
        
        \textbf{Model Inference:}
        \begin{itemize}[leftmargin=1.5em, label=\textbullet, nosep]
            \item $<$think$>$ : The query follows a standard movie search intent pattern. ``steven spielberg'' matches a director name commonly annotated in the dataset.$<$/think$>$ \textbf{\textcolor{decisionRed}{[Decision:DIRECT]}}
        \end{itemize}

        \textbf{Output :} $<$answer$>$(steven spielberg, DIRECT)$<$/answer$>$
    \end{tcolorbox}

    \caption{Different handling strategies for simple samples and difficult samples in \textbf{NE-R1}}
    \label{fig:casestudy}
\end{figure*}

\paragraph{False-positive and false-negative retrieval cases.}
Beyond successful examples, we also inspect failure modes of the learned retrieval policy. Figure~\ref{fig:fp_fn_cases} reports one false-positive retrieval case, where the model retrieves despite sufficient local evidence, and one false-negative case, where the model fails to retrieve for an ambiguous or knowledge-intensive mention. These cases clarify the remaining limitations of on-demand retrieval.

\begin{figure*}[!tbp]
\centering
\small
\sffamily
\begin{tcolorbox}[
    enhanced,
    title={\textbf{False Positive Retrieval: unnecessary search on an easy instance}},
    colframe=paperred,
    colback=DataPink,
    fonttitle=\bfseries,
    arc=2mm,
    boxrule=0.8pt,
    left=2.5mm, right=2.5mm, top=2.5mm, bottom=2.5mm,
    boxsep=1.2mm]
\textbf{Input:} ``\textit{find action movies with a lot of explosions}''

\textbf{Model trajectory:} $<$think$>$ The query mentions ``explosions'', which might refer to specific movie titles. External information may help identify relevant titles. $<$/think$>$ \textbf{[Decision: SEARCH]} $<$search$>$explosion action movies$<$/search$>$

\textbf{Retrieved information:} ``Top action movies featuring explosions: Die Hard, Speed, Mission Impossible...''

\textbf{Prediction:} \texttt{\{"Genre": ["action"], "Plot": ["explosions"]\}}

\textbf{Ground truth:} \texttt{\{"Genre": ["action"], "Plot": ["explosions"]\}}

\textbf{Analysis:} The retrieved results introduce many concrete movie titles, but the input only requires recognizing the genre and plot description. Retrieval is therefore redundant and may introduce noise, although it does not change the final prediction in this case.
\end{tcolorbox}

\begin{tcolorbox}[
    enhanced,
    title={\textbf{False Negative Retrieval: missed search on a hard instance}},
    colframe=TextOrange,
    colback=DataYellow,
    fonttitle=\bfseries,
    arc=2mm,
    boxrule=0.8pt,
    left=2.5mm, right=2.5mm, top=2.5mm, bottom=2.5mm,
    boxsep=1.2mm]
\textbf{Input:} ``\textit{I want to watch something like primer}''

\textbf{Model trajectory:} $<$think$>$ The word ``primer'' could refer to a common noun, such as a basic guide. The context is insufficient to confirm it as a movie title. $<$/think$>$ \textbf{[Decision: DIRECT]}

\textbf{Prediction:} misses the title entity ``primer''.

\textbf{Ground truth:} \texttt{\{"Title": ["primer"]\}}

\textbf{Analysis:} ``Primer'' is a long-tail entity referring to a 2004 low-budget independent science-fiction film. Since the model's parametric knowledge is insufficient to confidently identify it as a movie title, retrieval should have been triggered. The missed search leads to an entity-type recognition error.
\end{tcolorbox}
\caption{Representative false-positive and false-negative retrieval cases on MIT-Movie. The false-positive case shows redundant retrieval on an easy query, while the false-negative case shows a missed retrieval opportunity for a long-tail movie title.}
\label{fig:fp_fn_cases}
\end{figure*}

\end{document}